\documentclass[pmlr]{jmlr}

\RequirePackage{graphicx}
 \usepackage{booktabs}
\usepackage{longtable}
\makeatletter
\def\set@curr@file#1{\def\@curr@file{#1}} 
\makeatother
\usepackage[load-configurations=version-1]{siunitx} 

\usepackage{multirow}
\newcommand{\best}[1]{\textbf{#1}}
\newcommand{\second}[1]{\underline{#1}}
\newcommand{\method}{OC-Distill}
\usepackage{adjustbox}
\usepackage{float}

\theorembodyfont{\upshape}
\theoremheaderfont{\scshape}
\theorempostheader{:}
\theoremsep{\newline}

\jmlrproceedings{PMLR}{Proceedings of Machine Learning Research}
\jmlrvolume{340}
\jmlryear{2026}
\jmlrworkshop{Machine Learning for Healthcare}

\title[Ontology-aware CL with Cross-Modal Distillation for ICU Risk Prediction]{OC-Distill: Ontology-aware Contrastive Learning with Cross-Modal Distillation for ICU Risk Prediction}

\author{\Name{Zhongyuan Liang}
       \Email{zhongyuan\_liang@berkeley.edu}\\ 
       \addr Computational Precision Health\\
       UC Berkeley, UCSF\\
       \AND
       \Name{Junhyung Jo}
       \Email{junbro.jo@samsung.com}\\ 
       \addr Samsung Advanced Institute of Technology (SAIT)\\
       \AND
       \Name{Hyang-Jung Lee}
       \Email{hjeva.lee@samsung.com}\\ 
       \addr Samsung Advanced Institute of Technology (SAIT)\\
       \AND
       \Name{Sang Kyu Kim}
       \Email{sangq.kim@samsung.com}\\ 
       \addr Samsung Advanced Institute of Technology (SAIT)\\
       \AND
       \Name{Irene Y. Chen}
       \Email{iychen@berkeley.edu}\\ 
       \addr Computational Precision Health\\
       UC Berkeley, UCSF}

\begin{document}

\maketitle

\vspace{-.2in}
\begin{abstract}
Early prediction of severe clinical deterioration and remaining length of stay can enable timely intervention and better resource allocation in high-acuity settings such as the ICU. This has driven the development of machine learning models that leverage continuous streams of vital signs and other physiological signals for real-time risk prediction. Despite their promise, existing methods have important limitations. Contrastive pretraining treats all patients as equally strong negatives, failing to capture clinically meaningful similarity between patients with related diagnoses. Meanwhile, downstream fine-tuning typically ignores complementary modalities such as clinical notes, which provide rich contextual information unavailable in physiological signals alone. To address these challenges, we propose \method, a two-stage framework that leverages multimodal supervision during training while requiring only vital signs at inference. In the first stage, we introduce an ontology-aware contrastive objective that exploits the ICD hierarchy to quantify patient similarity and learn clinically grounded representations. In the second stage, we fine-tune the pretrained encoder via cross-modal knowledge distillation, transferring complementary information from clinical notes into the model. Across multiple ICU prediction tasks on MIMIC, \method\ demonstrates improved label efficiency and achieves state-of-the-art performance among methods that use only vital signs at inference.
\\ \textbf{Code:} \url{https://github.com/the-chen-lab/OC-Distill}
\end{abstract}

\section{Introduction}\label{sec: intro}
\begin{figure}[th]
    \centering
    \includegraphics[width=\textwidth]{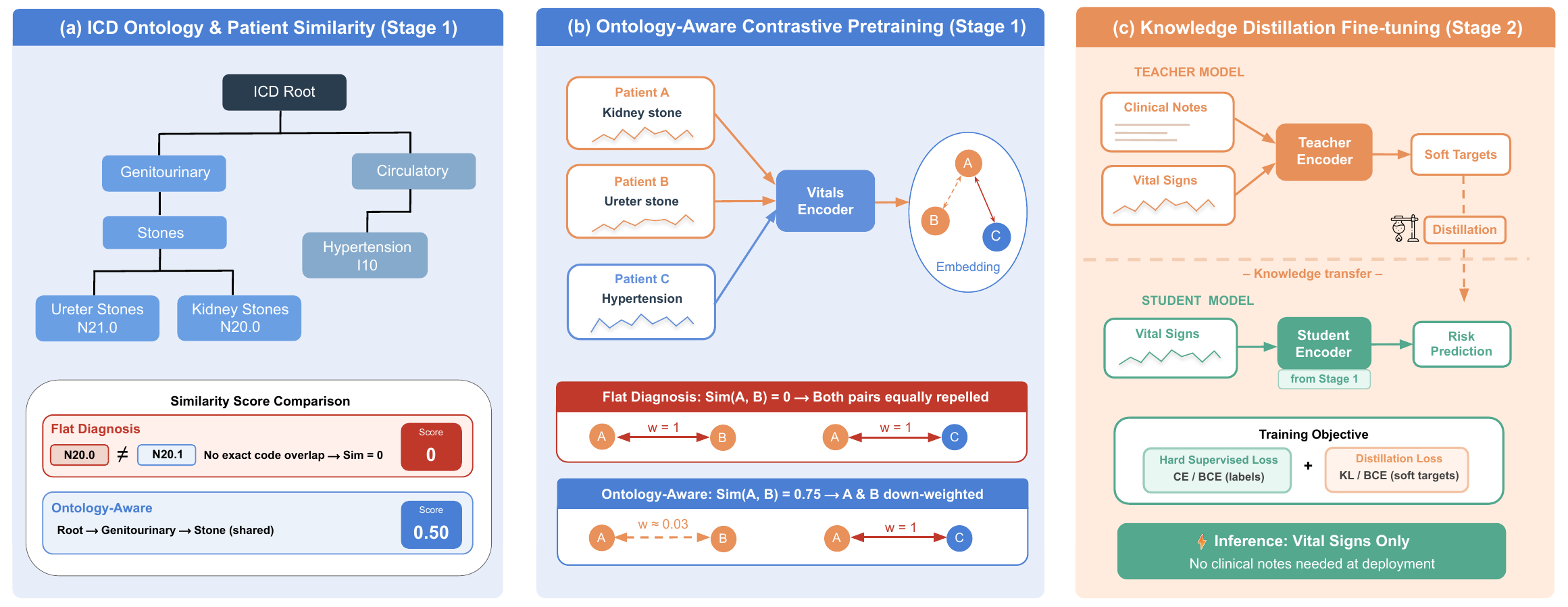}
    \vspace{-0.1in}
    \caption{Overview of our two-stage framework. (a) We compute patient similarity from the ICD diagnosis hierarchy, providing more informative scores than flat diagnosis matching. (b) Ontology-aware contrastive pretraining down-weights clinically similar negative pairs while strongly repelling dissimilar pairs. (c) Knowledge distillation finetuning transfers representations from a multimodal teacher to a vitals-only student, enabling inference without clinical notes at deployment.}
    \label{fig:diagram}
\end{figure}

Early prediction of clinical deterioration in the intensive care unit (ICU) is critical for timely intervention and effective resource allocation. The availability of continuous physiological monitoring has enabled the development of machine learning (ML) models that can identify patients at elevated risk of adverse outcomes by processing streams of vital signs and other bedside time-series data \citep{suresh2017clinical, purushotham2018benchmarking, wang2020mimic, hu2026protocol}. Despite substantial progress, current approaches face two key limitations that constrain their effectiveness in real-world ICU settings.

First, contrastive learning (CL) has shown promise for pretraining physiological signals, yet existing methods treat all patients as equally strong negatives during training \citep{chen2020simpleframeworkcontrastivelearning, raghu2023sequential, king2024efficient}, ignoring that patients often share related disease profiles. Recent work has attempted to address this by computing the percentage of exact ICD code matches between patients as a proxy for clinical similarity \citep{ding2024distilling}, but this flat matching approach overlooks the hierarchical structure of diagnosis taxonomies. Because the ICD taxonomy contains thousands of specific codes, exact matches are sparse even among clinically related patients. For example, a patient with ureter stones and a patient with kidney stones share substantial pathophysiological commonalities encoded in the ICD hierarchy, yet flat matching assigns them zero similarity. Consequently, pretraining strategies that ignore these hierarchical relationships may learn representations that fail to reflect clinical similarity.

Second, training on physiological time series alone ignores other rich modalities in the electronic health record (EHR). In particular, clinical notes contain clinical reasoning and contextual information, such as patient history and behavioral observations, that are not captured in physiological signals alone \citep{seinen2025using, pmlr-v287-liang25a, du2026possible}. While recent work explores multimodal fusion approaches \citep{shukla2021integratingphysiologicaltimeseries, wang2022integratingphysiologicaltimeseries, lyu2023multimodal, battula2024enhancing, meng2026clcr}, these methods require all modalities as inputs at inference time—an assumption that often fails in practice. For example, clinical notes are time-consuming to generate, whereas ICU workflows demand rapid, continuous monitoring. Furthermore, different modalities are often stored in disparate systems, making real-world deployment of such approaches impractical.

To address these limitations, we introduce \method~(\textbf{O}ntology-aware \textbf{C}ontrastive learning with cross-modal \textbf{Distill}ation), a two-stage framework for ICU risk prediction that leverages multimodal supervision during training while requiring only vital signs at inference (Fig~\ref{fig:diagram}). In the first stage, we develop an ontology-aware CL approach that incorporates patient similarity from the ICD diagnosis hierarchy. Rather than applying a uniform contrastive penalty to all negative pairs, we down-weight clinically similar patients during pretraining, allowing the model to learn representations that preserve the structure of disease relationships. In the second stage, we finetune the pretrained encoder via cross-modal knowledge distillation \citep{hinton2015distillingknowledgeneuralnetwork, lan2025acam, duan2026maventreinforcedheterogeneousdistillation, lan2026reco, ding2026synergistic}, where a teacher model trained on both vital signs and clinical notes transfers its representations to a student model initialized from the pretrained encoder that operates exclusively on vitals. This enables the student to benefit from clinical context during training without requiring multimodal inputs at deployment.

We evaluate our approach on MIMIC \citep{harutyunyan2019multitask, johnson2023mimic} across multiple ICU prediction tasks, label fractions, and observation horizons. Across these settings, incorporating similarity from the ICD ontology into contrastive pretraining yields more clinically informed representations, demonstrating strong label efficiency in both linear evaluation and end-to-end fine-tuning. Furthermore, finetuning with knowledge distillation from clinical notes provides additional gains, and the full \method\ pipeline achieves state-of-the-art performance among methods that operate on vital signs alone at inference time. In summary, we make the following contributions:

\begin{enumerate}
    \item An ontology-aware CL method that integrates patient similarity derived from the ICD ontology, yielding more clinically grounded and label-efficient representations.
    \item A multimodal distillation approach that distills knowledge from clinical notes into a vitals model, improving downstream performance without requiring multimodal inputs at inference.
    \item A comprehensive empirical evaluation on the MIMIC benchmarks, demonstrating state-of-the-art performance across tasks, label fractions, and observation horizons.
\end{enumerate}

\subsection*{Generalizable Insights about Machine Learning in the Context of Healthcare}
This work offers several insights for machine learning in healthcare beyond the specific task of ICU risk prediction. First, our results show that incorporating domain-specific clinical structure can meaningfully improve machine learning performance. In particular, diagnostic ontologies such as ICD provide a readily available and clinically meaningful source of supervision. When used in contrastive pretraining, this structure can substantially reduce the amount of labeled data needed to achieve strong downstream performance. Second, we identify a fundamental tension between current multimodal fusion approaches and the constraints of real-world clinical deployment. In practice, data modalities are distributed across separate systems, and some modalities, such as clinical notes, are time-consuming to generate. At the same time, ICU workflows require rapid and continuous monitoring. Our framework addresses this mismatch through knowledge distillation, allowing models to benefit from rich multimodal supervision during training while remaining practical and efficient at deployment. Overall, this work presents a generalizable approach for leveraging clinical domain structure in the learning pipeline while respecting the practical constraints of real-world deployment.

\section{Related Work}\label{sec: related}
\subsection{Contrastive Pretraining for Clinical Time Series}
CL is widely used for pretraining representations from unlabeled data due to its simple objective and strong empirical performance \citep{jaiswal2021surveycontrastiveselfsupervisedlearning, kiyasseh2021clocs, gopal20213kg}. A prominent example is SimCLR, which learns embeddings that pull together augmented views of the same instance while uniformly pushing apart other instances in the batch using the NT-Xent loss \citep{chen2020simpleframeworkcontrastivelearning}. Although originally developed for vision, CL has since been broadly adopted for representation learning on clinical time series \citep{weatherhead2022learning, li2022multi, zhang2022forecasting, raghu2023sequential, noroozizadeh2023temporal, liu2023self, king2024efficient}. However, existing methods treat all patients as equally strong negatives, ignoring that patients in clinical settings often share related diagnoses and disease profiles. While recent work has explored flat ICD code overlap as a similarity signal \citep{ding2024distilling}, this ignores the hierarchical structure of diagnosis taxonomies. We therefore leverage the full ICD ontology to derive fine-grained patient similarity and down-weight clinically similar negatives during contrastive pretraining, grounding the objective in clinically meaningful disease relationships.

\subsection{Multimodal Learning for Clinical Time Series}
Recent work shows that multimodal models can outperform unimodal baselines by leveraging complementary clinical signals \citep{khader2023medical, jorf2025medpatch}. Accordingly, many approaches incorporate fusion mechanisms and report improved performance when combining physiologic time series with additional modalities, such as clinical notes \citep{lyu2023multimodal, battula2024enhancing, hooman2025equitableelectronichealthrecord}. While effective, fusion-based approaches require all modalities to be available as inputs at deployment, an assumption that often fails in practice \citep{lang2025retrieval, lang2025redeeming, fu2026missing}. We instead study \emph{training-time} multimodal supervision, leveraging auxiliary modality (e.g., clinical notes) during training while using only real-time vital signs at deployment for rapid prediction.

Within this paradigm, \citet{ketabi2025bridging} adopt CLIP-style contrastive pretraining to align physiologic signals with clinical notes, and subsequently fine-tune the resulting time-series encoder for downstream tasks. \citet{ding2024distilling} additionally regularize the learned representations by using a cross-entropy loss to match a diagnosis-based similarity distribution, alongside the CLIP objective. Alternative methods draw on privileged-information learning \citep{vapnik2009new, chen2021learning}, particularly modality synthesis, where auxiliary modalities available during training are generated at inference time to support prediction \citep{chartsias2017multimodal, yu20183d}.
\section{Methods}\label{sec: methodology}
We define a two-stage framework consisting of ontology-aware contrastive pretraining followed by knowledge distillation fine-tuning. In the first stage (Section \ref{sec:stage1}), we learn representations from vital-sign time series using CL that incorporates similarity scores derived from the ICD ontology. In the second stage (Section \ref{sec:stage2}), we fine-tune the model using a knowledge distillation framework, where representations learned from clinical notes are distilled into the time-series encoder for downstream ICU prediction tasks.

\subsection{Stage 1: Ontology-aware Contrastive Pretraining}
\label{sec:stage1}
Standard CL frameworks treat all negative pairs equally during training \citep{chen2020simpleframeworkcontrastivelearning}. However, in clinical settings, patients often share related disease profiles and should not be pushed arbitrarily far apart in the representation space. To address this limitation, we propose an ontology-aware CL approach that uses ICD-derived patient similarity to down-weight clinically similar negatives during pretraining. We begin by defining a measure of patient similarity based on the hierarchical structure of ICD ontology.

\subsubsection{Patient Similarity Score Derivation}
\label{sec:similarity}
We obtain the ICD ontology from BioPortal \citep{noy2009bioportal} and use it to construct a hierarchical representation of diagnosis concepts. The resulting structure is a rooted tree: diagnoses share a common root and branch into progressively more specific categories (e.g., digestive diseases, circulatory diseases, congenital anomalies), with leaf nodes corresponding to individual ICD codes. The tree has maximum depth six and includes 14{,}870 nodes.

For two ICD codes $a$ and $b$, let $\mathrm{Path}(a)$ denote the set of nodes on the path from the root to $a$, $\mathrm{Path}(b)$ denote the set of nodes on the path from the root to $b$, and define $\mathrm{Shared}(a,b)=|\mathrm{Path}(a)\cap \mathrm{Path}(b)|-1$ as the number of nodes the two paths have in common excluding the root. We then define the similarity between $a$ and $b$ as the Jaccard overlap of their root-excluded path sets:

\begin{equation*}
s(a,b) \;=\; \frac{\mathrm{Shared}(a,b)}{|\mathrm{Path}(a)\cup \mathrm{Path}(b)|-1},
\end{equation*}

\noindent where $|\cdot|$ denotes set cardinality. This yields a similarity score that ranges from 0 (only the root is shared) to 1 (identical codes).

Patients typically have multiple diagnoses. For two patients $A$ and $B$ with diagnosis sets
$A=\{a_1,\ldots,a_m\}$ and $B=\{b_1,\ldots,b_n\}$, we compute patient-level similarity via a symmetric
\emph{best-match} aggregation over codes. Specifically, for each code in $A$, we take its maximum ontology similarity to any code in $B$ and average across codes, and we repeat the same procedure in the reverse direction for $B$:
\begin{equation*}
\text{best}_A(a_i) = \max_{b \in B} s(a_i, b), \quad
\text{avg}_A = \frac{1}{m} \sum_{i=1}^{m} \text{best}_A(a_i).
\end{equation*}
\begin{equation*}
\text{best}_B(b_j) = \max_{a \in A} s(a, b_j), \quad
\text{avg}_B = \frac{1}{n} \sum_{j=1}^{n} \text{best}_B(b_j).
\end{equation*}

We then define the overall patient similarity as the mean of the two averages:

\begin{equation*}
\text{Sim}(A, B) = \frac{1}{2}\left(\text{avg}_A + \text{avg}_B\right).
\end{equation*}

\subsubsection{Ontology-aware Contrastive Loss}
We next use the derived similarity score to guide CL on vital-sign time series. Let $\mathbf{e}_i$ denote the $\ell_2$-normalized representation of patient $i$ produced by a time-series encoder $f_{\text{vitals}}$. Following standard CL practice, for each patient $i$ in a batch, we construct a positive example $\mathbf{e}_i^+$ for each anchor $\mathbf{e}_i$ by applying transformations to the input. The representations of all other patients in the same batch are treated as negatives. The standard NT-Xent loss pushes all negative samples away from the anchor with equal force:

\begin{equation*}
\mathcal{L}_{\text{NT-Xent}} = -\log 
\frac{\exp(\mathbf{e}_i^\top \mathbf{e}_i^+ / \tau)}
{\exp(\mathbf{e}_i^\top \mathbf{e}_i^+ / \tau) + \sum_{j \neq i} \exp(\mathbf{e}_i^\top \mathbf{e}_j / \tau)},
\end{equation*}

\noindent where $\tau$ is a temperature parameter.

We then incorporate the patient similarity score from Section \ref{sec:similarity} into the contrastive loss. Intuitively, patients with highly overlapping diagnosis profiles should not be treated as equally strong negatives, since forcing their representations far apart can distort the embedding space. Concretely, we propose a weighted contrastive objective in which each negative pair $(i,j)$ is assigned a weight
$w_{ij}=\Phi(\mathrm{Sim}(A_i,A_j))\in[0,1]$. 

\begin{equation*}
\mathcal{L}_{\text{OW-NTXent}}^{(i)} = -\log 
\frac{\exp(\mathbf{e}_i^\top \mathbf{e}_i^+ / \tau)}
{\exp(\mathbf{e}_i^\top \mathbf{e}_i^+ / \tau) + \sum_{j \neq i} w_{ij}\,\exp(\mathbf{e}_i^\top \mathbf{e}_j / \tau)}.
\end{equation*}

The transformation $\Phi:[0,1]\rightarrow[0,1]$ is a monotone decreasing transformation that controls the sharpness of down-weighting and can be instantiated by different parametric forms, including a power transform $\Phi(s)=(1-s)^{\gamma}$, an exponential decay $\Phi(s)=\exp(-\gamma s)$, or a hard threshold $\Phi(s)=\mathbb{I}[s<\delta]$. Larger $\gamma$ or a smaller threshold $\delta$ yields a sharper weighting, more aggressively down-weighting highly similar pairs while placing relatively greater weight on dissimilar negatives.

\subsection{Stage 2: Knowledge Distillation Finetuning}
\label{sec:stage2}
We next finetune the pretrained time series encoder using knowledge distillation to transfer complementary information from clinical notes. We first train a teacher model that uses clinical notes together with vital signs to generate informative soft targets (Section \ref{subsec: teacher}). We then finetune a student model initialized from the stage 1 pretrained encoder by distilling the teacher outputs while also optimizing the standard supervised loss (Section \ref{subsec: student}).

\subsubsection{Teacher Model Training}
\label{subsec: teacher}
Let $\{(\mathbf{x}_i, y_i)\}_{i=1}^{n}$ denote the training set, where
$\mathbf{x}_i = [\mathbf{x}_i^{(orig)}, \mathbf{x}_i^{(vitals)}]$.
Here, $\mathbf{x}_i^{(orig)}$ is the raw clinical note sequence and
$\mathbf{x}_i^{(vitals)} \in \mathbb{R}^{T \times c}$ denotes the time-series vital signs with $c$ channels over $T$ time steps. The label $y_i$ is the outcome of interest.

Clinical notes are often long and sparse, mixing salient information with noise. LLMs have demonstrated the ability to reason over complex information and generate concise summaries for downstream tasks \citep{leong2024gen, zhu2025ants, li2025enrich, lan2026alternating, guo2026toward, 11393831, yuan2026sp}. Their capabilities have also been explored across a broad range of technical and biomedical applications \citep{huang2024building, leong2024efficient, zeng2025FSDrive, nie2026understanding, li2026same, hu2026toward, zhang2026performance, yao2026measuring}. Motivated by \citet{battula2024enhancing}, we augment the raw notes with LLM-generated summaries. Specifically, we use GPT-4o in a zero-shot setting to generate a concise summary $\mathbf{x}_i^{(summ)}$ for each raw note $\mathbf{x}_i^{(orig)}$. During training, we randomly select between the raw note and its summary with probability $p$ and $1-p$, respectively:
\begin{equation*}
\mathbf{x}_i^{(notes)} =
\begin{cases}
\mathbf{x}_i^{(orig)} & \text{with probability } p,\\
\mathbf{x}_i^{(summ)} & \text{with probability } 1-p,
\end{cases}
\end{equation*}

For each example $i$, we encode each modality into a shared $d$-dimensional embedding space ($d=768$) using modality-specific encoders:
\begin{equation*}
\mathbf{h}_i^{(\text{notes})} = f_{\text{notes}}\!\left(\mathbf{x}_i^{(\text{notes})}\right) \in \mathbb{R}^{d},
\qquad
\mathbf{h}_i^{(\text{vitals})} = f_{\text{vitals}}\!\left(\mathbf{x}_i^{(\text{vitals})}\right) \in \mathbb{R}^{d}.
\end{equation*}

We then fuse the modalities by element-wise addition:
\begin{equation*}
\mathbf{h}_i^{(\text{fuse})} = \mathbf{h}_i^{(\text{notes})} + \mathbf{h}_i^{(\text{vitals})}.
\end{equation*}

The fused representation is passed to a classification head to produce teacher logits
$\mathbf{z}_i^{(\text{teacher})}=g_{\text{teacher}}\!\left(\mathbf{h}_i^{(\text{fuse})}\right)$, followed by a softmax (multiclass) or sigmoid (binary) to obtain predicted probabilities $\hat{\mathbf{y}}_i^{(\text{teacher})}$. 

We jointly train the encoders end-to-end using a hard supervised loss $\mathcal{L}_{\text{hard}}^{(i)}$ on ground truth labels. We use cross-entropy (CE) loss for multi-class outcomes and binary cross-entropy (BCE) for binary outcomes.

\begin{equation*}
\begin{gathered}
\mathcal{L}_{\text{hard}}^{(i)} =
\begin{cases}
\mathrm{CE}\!\left(\hat{\mathbf{y}}_i^{(\text{teacher})}, y_i\right), & \text{multi-class},\\
\mathrm{BCE}\!\left(\hat{\mathbf{y}}_i^{(\text{teacher})}, y_i\right), & \text{binary},
\end{cases}
\qquad
\mathcal{L}_{\text{teacher}} =
\frac{1}{n}\sum_{i=1}^{n} \mathcal{L}_{\text{hard}}^{(i)}.
\end{gathered}
\end{equation*}

\subsubsection{Student Model Training}
\label{subsec: student}
After training the teacher model, we freeze its parameters and distill its learned knowledge into a student model that uses only time-series vital signs as input. For each example $i$, the student encoder produces a latent representation
\begin{equation*}
\mathbf{h}_i^{(\text{student})} = f_{\text{vitals}}\!\left(\mathbf{x}_i^{(\text{vitals})}\right),
\end{equation*}
which is then passed to a classification head to produce student logits $\mathbf{z}_i^{(\text{student})} = g_{\text{student}}\!\left(\mathbf{h}_i^{(\text{student})}\right).$

The student is trained with the same hard supervised loss $\mathcal{L}_{\text{hard}}^{(i)}$ and an additional soft distillation loss. For multi-class outcomes, we match predictions via KL-divergence:

\begin{equation*}
\begin{gathered}
\mathbf{p}_{i,T}^{(\text{teacher})}
= \mathrm{softmax}\!\left(\mathbf{z}_i^{(\text{teacher})}/T\right), \qquad
\mathbf{p}_{i,T}^{(\text{student})}
= \mathrm{softmax}\!\left(\mathbf{z}_i^{(\text{student})}/T\right), \\
\mathcal{L}_{\text{distill}}^{(i)}
= \mathrm{KL}\!\left(
\mathbf{p}_{i,T}^{(\text{teacher})}
\,\|\, 
\mathbf{p}_{i,T}^{(\text{student})}
\right).
\end{gathered}
\end{equation*}

For binary outcomes, we use the teacher’s sigmoid output as a soft target:

\begin{equation*}
\begin{gathered}
p_{i,T}^{(\text{teacher})}
= \sigma\!\left(z_i^{(\text{teacher})}/T\right), \quad
p_{i,T}^{(\text{student})}
= \sigma\!\left(z_i^{(\text{student})}/T\right), \\
\mathcal{L}_{\text{distill}}^{(i)}
= \mathrm{BCE}\!\left(
p_{i,T}^{(\text{student})},\,
p_{i,T}^{(\text{teacher})}
\right),
\end{gathered}
\end{equation*}

\noindent where $T$ is a temperature parameter that softens the probability distributions, allowing the student to learn from the teacher's uncertainty.

The final objective is
\begin{equation*}
\mathcal{L}_{\text{student}} = \frac{1}{n}\sum_{i=1}^{n}\left(\mathcal{L}_{\text{hard}}^{(i)} + \lambda_{\text{distill}}\,\mathcal{L}_{\text{distill}}^{(i)}\right),
\end{equation*}

\noindent where $\lambda_{\text{distill}}$ is a hyperparameter controlling distillation strength. After training, the student model is deployed using only vital signs as input, enabling real-time predictions without requiring clinical notes.
\section{Cohort and Experiment Setup}\label{sec: experiments}

\subsection{Cohort Selection and Benchmark Tasks}
For our experiments, we construct time-series features for each ICU stay from the MIMIC-III benchmark dataset \citep{harutyunyan2019multitask} and align them with the corresponding clinical notes \citep{johnson2016mimic}. We benchmark model performance on the following tasks widely used to evaluate ICU risk prediction models \citep{purushotham2018benchmarking, harutyunyan2019multitask, wang2020mimic}.
\begin{enumerate}
    \item \textbf{In-hospital mortality prediction}: A binary classification task that predicts in-hospital mortality after the first $T$ hours spent in the ICU. This task assesses the model's ability to identify high-risk patients early. We evaluate this task using the AUROC and the AUPRC.
    
    \item \textbf{Length of stay prediction}: A 10-class classification task that predicts the remaining ICU length of stay from the first $T$ hours of an ICU admission. Accurate length of stay prediction is important for hospital resource planning. We evaluate this task using the macro-averaged one-vs-rest AUROC and AUPRC.  
\end{enumerate}

For both tasks, we report results for $T=48$ hours in Section \ref{sec:results} and additionally evaluate $T \in \{72, 96\}$ hours in Appendix \ref{apd:first}, where we observe similar performance trends.

\subsection{Data Extraction and Feature Choices}
We follow the pre-processing pipeline of prior work \citep{harutyunyan2019multitask} and use 12 vital signs and physiological variables (diastolic blood pressure, fraction of inspired oxygen, glucose, heart rate, height, mean blood pressure, oxygen saturation, respiratory rate, systolic blood pressure, temperature, weight, and pH). For all tasks, we discretize the time series into one-hour intervals and concatenate a binary missingness indicator for each variable, yielding a time-series representation $\mathbf{x}_i^{(\mathrm{vitals})} \in \mathbb{R}^{T \times 24}$.

We then align clinical notes to each ICU stay and retain only those recorded within the first $T$ hours of the stay. We focus on nursing and radiology notes, the two most frequently available note types. For each stay, we sort notes by chart time and concatenate them into a single document. We then preprocess the text by lowercasing, removing bracketed de-identification spans, collapsing repeated separator characters, and normalizing whitespace. We additionally use GPT-4o to generate concise summaries of the clinical notes for data augmentation during teacher training. We accessed GPT-4o via the Microsoft Azure API, which is HIPAA-compliant and recommended by PhysioNet for use with MIMIC data. Appendix \ref{apd:teacher_summary_frac} reports ablation results comparing performance with and without LLM-generated summaries, showing consistent gains when summaries are combined with raw notes. Appendix~\ref{app:teacher_calibration} further shows that the multimodal teacher produces better-calibrated soft probabilities than a note-free teacher, motivating the use of distillation.

We split the dataset using the same train/test split as in \citet{harutyunyan2019multitask}. We report final performance on the test set and compute 95\% confidence intervals using 1{,}000 bootstrap resamples \citep{efron1994introduction}. Table~\ref{tab:data_size} summarizes the dataset sizes for each observation horizon $T$.

\begin{table}[t]
\centering
\small
\setlength{\tabcolsep}{10pt}
\renewcommand{\arraystretch}{1.15}
\begin{tabular}{l c c}
\toprule
\textbf{Observation horizon $T$} & \textbf{Training} & \textbf{Test} \\
\midrule
48 hours & 16{,}861 & 3{,}055 \\
72 hours & 11{,}003 & 1{,}958 \\
96 hours & 7{,}788  & 1{,}418 \\
\bottomrule
\end{tabular}
\caption{Dataset sizes (ICU stays) in the constructed cohort for each observation horizon.}
\label{tab:data_size}
\end{table}

\subsection{Dataset Choice and External Validation}
Our experiments are conducted on MIMIC-III, which uniquely provides longitudinal clinical notes, specifically nursing notes written throughout a patient's ICU stay. Other commonly used benchmarks such as MIMIC-IV \citep{johnson2023mimic} include only discharge summaries and radiology reports, making them less suitable for our setting where notes serve as a rich source of training-time supervision. To assess generalizability beyond MIMIC-III, we provide additional evaluation of our method on MIMIC-IV vital signs in Appendix~\ref{apd:mimic4} and eICU vital signs \citep{pollard2018eicu} in Appendix~\ref{apd:eicu}, where \method\ continues to show strong performance, demonstrating robustness across datasets.

\subsection{Model Training and Implementation Details}
To ensure a fair comparison, we use consistent model architectures across all methods and baselines.
We parameterize the vitals encoder $f_{\text{vitals}}$ as a BERT-style transformer \citep{devlin2019bertpretrainingdeepbidirectional} with two layers and four attention heads. 
For clinical notes, we parameterize $f_{\text{notes}}$ using Bio\_ClinicalBERT \citep{alsentzer2019publicly}, a transformer-based language model shown to capture the terminology and structure of clinical documentation effectively.

For contrastive pretraining, positive pairs are constructed by applying transformations to the same vitals sequence. Specifically, given an input time series, we independently generate two views using three augmentations: (1) \emph{Gaussian jitter}, which adds zero-mean Gaussian noise to observed vital values; (2) \emph{random time masking}, which masks random timesteps by setting all features at those timesteps to zero; and (3) \emph{random feature masking}, which masks random vital-sign channels across all timesteps by setting their values to zero. We train with learning rate $\eta = 10^{-4}$, temperature $\tau = 1.0$, and batch size of $4096$ for $1000$ epochs. We use the power transformation $\Phi(s)=(1-s)^{\gamma}$ with $\gamma=5$ as the similarity-weight transformation, and we study alternative choices of $\gamma$ values and $\Phi$ in Section \ref{sec:ablation_study}.

For knowledge distillation fine-tuning, we perform hyperparameter tuning via grid search over learning rate $\eta \in \{1\times10^{-4}, 5\times10^{-4}, 5\times10^{-5}\}$, distillation temperature $T \in \{1,2,5\}$, distillation loss weight $\lambda_{\text{distill}} \in \{1,5,10\}$, and the probability of using an LLM-generated note summary during training $p \in \{0,0.5,1.0\}$. Each model is trained for 100 epochs, and for each configuration we monitor validation AUROC throughout training and select the checkpoint with the best validation performance. The deployed student model contains approximately 19.3M parameters. On a single NVIDIA L40S GPU, contrastive pretraining takes 1.8 hours, the distillation pipeline takes 2.4 hours, and inference takes 0.0004 seconds per sample. All experiments were run on a single NVIDIA L40S GPU.
\section{Results} \label{sec:results}

In this section, we report results of \method\ and compare against a range of baselines. In Section \ref{sec:contrastive_results}, we evaluate ontology-aware contrastive pretraining, analyzing how ontology-derived similarities shape the embedding space and improve label efficiency. In Section \ref{sec:full_pipeline_results}, we evaluate the complete \method\ pipeline with distillation finetuning and show additional gains from incorporating clinical notes.

\subsection{Contrastive Pretraining Results}
\label{sec:contrastive_results}
\textbf{Baselines.}
To quantify the benefit of leveraging the ICD ontology during contrastive~pretraining, we compare our method against two baselines. (a) \textit{SimCLR} \citep{chen2020simpleframeworkcontrastivelearning} uses the standard NT-Xent objective and treats all non-matching instances as equally strong negatives (i.e., uniform weights $w_{ij}=1$). (b) \textit{Flat Diagnosis Match} \citep{ding2024distilling} defines patient similarity via exact ICD-code overlap, computed as the number of shared ICD codes normalized by the total number of codes across the two patients. We then plug this flat similarity into the same reweighting formulation as our negative-weighted loss function, so any performance differences isolate the contribution of exploiting the ICD ontology.

\subsubsection{Similarity Analysis and Representation Geometry}
\label{sec:similarity_analysis}
\begin{figure}[H]
    \centering
    \includegraphics[width=0.9\linewidth]{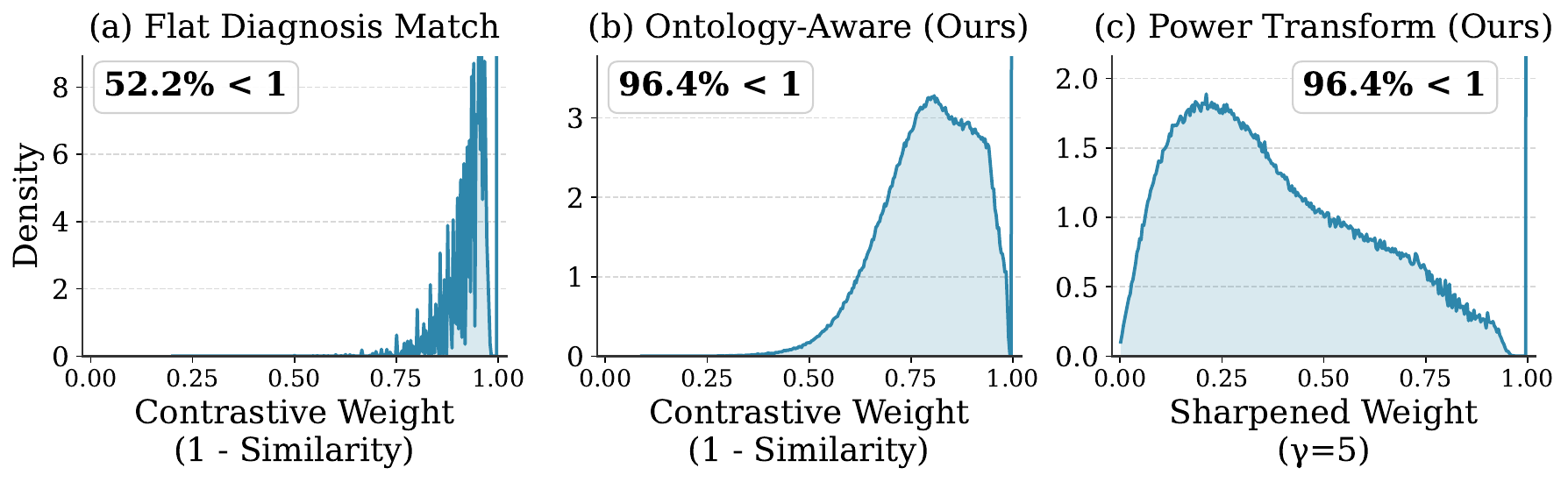}
    \caption{Distribution of contrastive weights for patient pairs. Flat diagnosis matching yields zero similarity ($\text{weight}=1$) for half of pairs. Ontology-aware similarity produces a smoother distribution with gradations, while a power transformation sharpens the weights. Percentages indicate the fraction with weight $< 1$.}
    \label{fig:sim_distribution}
\end{figure}

\textbf{Similarity Analysis.} We begin by examining how different similarity definitions translate into pairwise contrastive weights. As shown in Figure~\ref{fig:sim_distribution}, flat diagnosis matching is inherently coarse. With thousands of ICD codes, exact matching overlooks the hierarchy and assigns zero similarity ($\text{weight}=1$) to nearly half of patient pairs, effectively treating a large portion of negatives as equally dissimilar. In contrast, ontology-aware similarity yields a more informative measure that distributes weight across a continuum, with fewer than 4\% of pairs assigned zero similarity. A power transformation of the ontology-aware similarity further sharpens this spectrum, shifting more weights toward 0 and enabling more aggressive reweighting during pretraining.

\vspace{0.1in}

\noindent\textbf{Embedding-space similarity analysis.} We next evaluate whether the pretrained encoder produces embeddings that reflect the ontology-derived patient similarity structure.
\vspace{-0.28in}
\begin{figure}[H]
\begin{minipage}{0.48\textwidth}
By incorporating ontology-aware weighting into the contrastive loss objective, patients that are close in embedding space should also be clinically similar in terms of diagnosis structure. We test this by comparing diagnosis similarity between nearest-neighbor pairs ($K$=5) in the learned embedding space and randomly sampled pairs. As shown in Figure~\ref{fig:embed_similarity}, nearest-neighbor pairs exhibit significantly higher diagnosis similarity than random pairs (Mann--Whitney $U$: $p<$0.0001; $r$=0.198), validating that the representations capture diagnosis-relevant information. Additional results for different neighborhood sizes $K$ are reported in Appendix~\ref{apd:knn_sensitivity}, confirming that the conclusion is robust to the choice of $K$.

\end{minipage}
\enspace
\begin{minipage}{0.51\linewidth}
    \includegraphics[width=\linewidth]{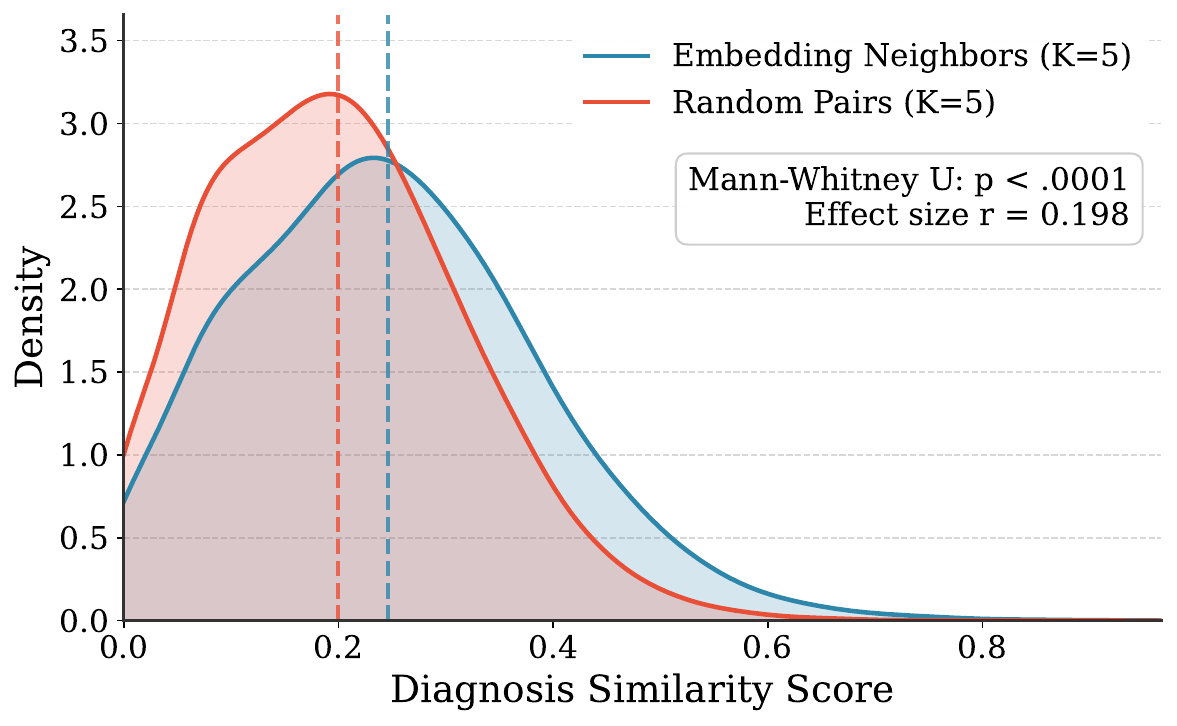}
    \vspace{-.3in}
    \caption{Diagnosis similarity distributions for embedding neighbors vs. random pairs. Embedding neighbors show higher similarity.}
    \label{fig:embed_similarity}
\end{minipage}
\end{figure}

\subsubsection{Pretraining Performance}
\label{sec:linear_probe_eval}

\begin{table}[t]
\centering
\normalsize  
\setlength{\tabcolsep}{6pt} 
\renewcommand{\arraystretch}{1.15}

\begin{adjustbox}{max width=\linewidth}
\begin{tabular}{l c cc cc}
\toprule
& & \multicolumn{2}{c}{\textbf{In-Hospital Mortality}} & \multicolumn{2}{c}{\textbf{Length of Stay}} \\
\cmidrule(lr){3-4}\cmidrule(lr){5-6}
\textbf{Methods} & \textbf{Labels} & \textbf{AUROC} & \textbf{AUPRC} & \textbf{AUROC} & \textbf{AUPRC} \\
\midrule

\multirow{3}{*}{SimCLR}
          & 1\%  & 0.635 (0.603--0.666) & 0.201 (0.172--0.238) & 0.533 (0.519--0.547) & 0.122 (0.119--0.130) \\
          & 5\%  & 0.723 (0.695--0.750) & 0.279 (0.235--0.328) & 0.588 (0.572--0.602) & 0.142 (0.138--0.151) \\
          & 10\% & 0.728 (0.701--0.754) & 0.289 (0.244--0.340) & 0.610 (0.595--0.624) & 0.149 (0.144--0.158) \\
\midrule

\multirow{3}{*}{Flat Diagnosis CL}
          & 1\%  & \second{0.647 (0.616--0.677)} & \second{0.208 (0.177--0.247)} & \second{0.536 (0.521--0.550)} & \best{0.123 (0.120--0.131)} \\
          & 5\%  & \second{0.731 (0.703--0.756)} & \second{0.294 (0.250--0.346)} & \second{0.591 (0.576--0.605)} & \second{0.142 (0.139--0.151)} \\
          & 10\% & \second{0.741 (0.714--0.765)} & \second{0.301 (0.257--0.352)} & \second{0.610 (0.596--0.624)} & \second{0.150 (0.144--0.159)} \\
\midrule

\multirow{3}{*}{Ontology-Aware CL (Ours)}
          & 1\%  & \best{0.673 (0.643--0.702)} & \best{0.230 (0.196--0.274)} & \best{0.541 (0.525--0.555)} & \best{0.123 (0.120--0.131)} \\
          & 5\%  & \best{0.750 (0.723--0.775)} & \best{0.319 (0.271--0.372)} & \best{0.600 (0.585--0.614)} & \best{0.144 (0.140--0.154)} \\
          & 10\% & \best{0.757 (0.731--0.781)} & \best{0.328 (0.278--0.382)} & \best{0.616 (0.602--0.629)} & \best{0.152 (0.147--0.161)} \\
\bottomrule
\end{tabular}
\end{adjustbox}

\caption{Linear evaluation of contrastive pretraining methods with 1\%, 5\%, and 10\% labeled training data. Ontology-aware CL consistently achieves the best results. Best is \textbf{bold}, second-best is \underline{underlined}.}
\label{tab:linear_prob}
\end{table}

\begin{table}[t]
\centering
\normalsize  
\setlength{\tabcolsep}{6pt}
\renewcommand{\arraystretch}{1.15}

\begin{adjustbox}{max width=\linewidth}
\begin{tabular}{l c cc cc}
\toprule
& & \multicolumn{2}{c}{\textbf{In-Hospital Mortality}} & \multicolumn{2}{c}{\textbf{Length of Stay}} \\
\cmidrule(lr){3-4}\cmidrule(lr){5-6}
\textbf{Methods} & \textbf{Labels} & \textbf{AUROC} & \textbf{AUPRC} & \textbf{AUROC} & \textbf{AUPRC} \\
\midrule

\multirow{2}{*}{SimCLR}
          & 50\%  & 0.730 (0.699--0.757) & 0.332 (0.281--0.386) & 0.668 (0.656--0.680) & 0.163 (0.159--0.173) \\
          & 100\% & 0.781 (0.755--0.805) & 0.377 (0.323--0.431) & \second{0.676 (0.664--0.688)} & \second{0.170 (0.166--0.180)} \\
\midrule

\multirow{2}{*}{Flat Diagnosis CL}
          & 50\%  & \second{0.763 (0.735--0.788)} & \best{0.360 (0.312--0.413)} & \second{0.669 (0.657--0.682)} & \second{0.164 (0.160--0.174)} \\
          & 100\% & \second{0.786 (0.760--0.809)} & \best{0.387 (0.333--0.439)} & \best{0.678 (0.666--0.690)} & \second{0.170 (0.166--0.180)} \\
\midrule

\multirow{2}{*}{Ontology-Aware CL (Ours)}
          & 50\%  & \best{0.778 (0.752--0.802)} & \second{0.359 (0.307--0.412)} & \best{0.673 (0.659--0.685)} & \best{0.165 (0.161--0.175)} \\
          & 100\% & \best{0.789 (0.764--0.812)} & \second{0.379 (0.326--0.434)} & \second{0.676 (0.663--0.688)} & \best{0.173 (0.168--0.184)} \\
\bottomrule
\end{tabular}
\end{adjustbox}

\caption{Full fine-tuning evaluation of contrastive pretraining models. Ontology-aware CL remains competitive and delivers strong performance across tasks. Best is \textbf{bold}, second-best is \underline{underlined}.}
\label{tab:fullfinetune}
\end{table}

\textbf{Linear evaluation.} We first assess the effect of contrastive pretraining via linear probing, where we freeze the encoder and train a linear classifier on top of the learned embeddings. Linear probing is most informative under label scarcity, and following standard protocol, we report results using 1\%, 5\%, and 10\% labeled training data (Table~\ref{tab:linear_prob}). Compared to SimCLR, Flat Diagnosis CL consistently yields stronger performance, highlighting the value of incorporating diagnosis supervision beyond augmentations alone. Ontology-Aware CL further improves upon Flat Diagnosis CL and achieves the best results across tasks and all label fractions, with the largest gains in the low-label regime. These findings indicate that leveraging ICD ontology structure produces more label-efficient representations.
\begin{figure}[H]
\begin{minipage}{0.48\textwidth}
\textbf{Cross-time generalization.} We further evaluate whether pretraining on $T=48$ hour time-series features yields representations that generalize to downstream tasks with longer observation windows. Figure~\ref{fig:cross_time} reports linear probing performance evaluated at downstream horizons of $T=48,\ 72,$ and $96$ hours. As expected, performance decreases for all methods as the downstream horizon increases. However, Ontology-Aware CL consistently remains the top-performing approach across observation horizons, demonstrating more robust cross-time generalization.
\end{minipage}
\enspace
\begin{minipage}{0.51\linewidth}
    \includegraphics[width=\linewidth]{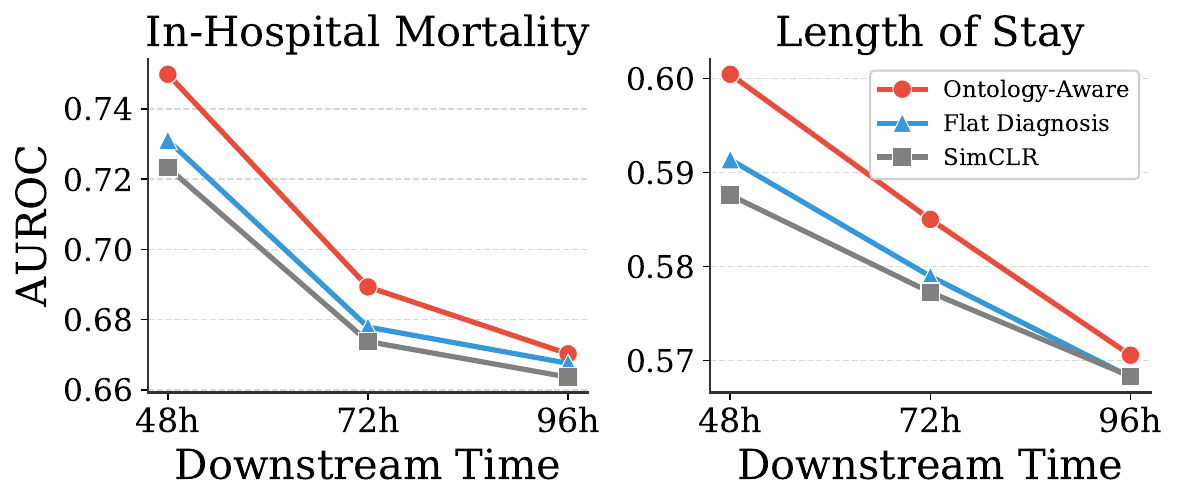}
    \vspace{-.3in}
    \caption{Cross-time generalization of linear-probe performance. Models were pretrained~on 48-hour windows and evaluated at downstream horizons of 48, 72, and 96 hours.}
    \label{fig:cross_time}
\end{minipage}
\end{figure}

\noindent \textbf{Full fine-tuning.} We lastly evaluate pretraining performance under full fine-tuning. In this setting, we initialize the model from pretrained weights and update all parameters during supervised training. Following standard protocol, we evaluate with 50\% and 100\% labeled data, as full fine-tuning requires sufficient labeled examples to update all parameters reliably. Table~\ref{tab:fullfinetune} reports results, where Ontology-Aware CL remains strong across tasks.

\subsubsection{Ablation Studies}
\label{sec:ablation_study}

\begin{figure}[t]
    \centering
    \includegraphics[width=\linewidth]{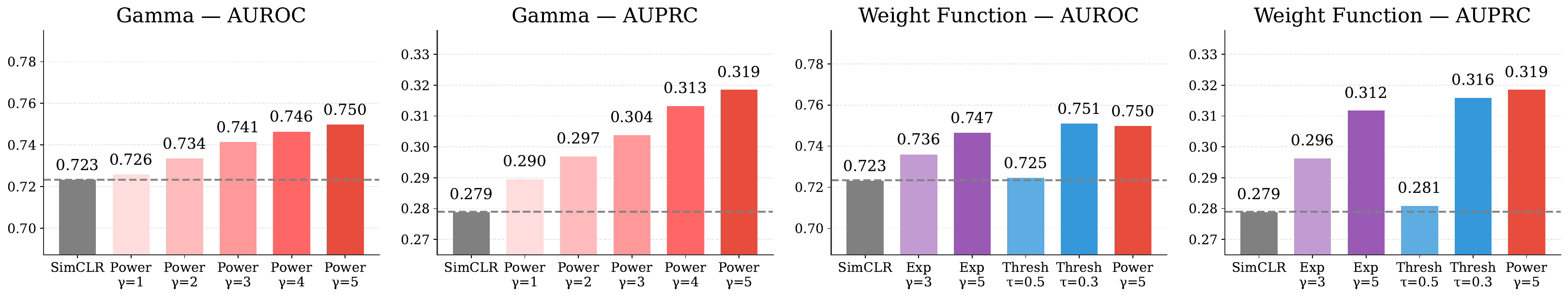}
    \vspace{-0.15in}
    \caption{Linear probing performance under different weight transformations, evaluated using 5\% labeled training data. Left pair: effect of the power-transform exponent $\gamma$. Right pair: comparison across weight functions. Similar trends are observed across other labeled-data fractions.}
    \label{fig:transformation}
\end{figure}

\textbf{Ablation effects of weight transformations.}
Our experiments incorporate similarity scores into the contrastive loss using the power transform $\Phi(s)=(1-s)^{\gamma}$ with $\gamma=5$. We ablate this design by varying $\gamma$ and the transformation family to quantify the effect of the transformation choice on performance. Figure~\ref{fig:transformation} reports linear probing results under these settings. We observe that all transformations consistently outperform SimCLR. Within the power family, performance improves as $\gamma$ increases, suggesting that more aggressive reweighting improves discriminability in supervised tasks. Exponential and threshold transforms show similar patterns, with sharpened settings (e.g., $\gamma=5$, $\tau=0.3$) achieving the best performance. Together, these results highlight the benefit of emphasizing clinically meaningful similarity structure during pretraining.

We further compare our simple Jaccard path-overlap similarity against established ontology similarity measures in Appendix~\ref{app:ontology_similarity_metrics}. Results show that our metric is strongly correlated with existing ontology similarity measures \citep{rada1989development, resnik1995using, jiang1997semantic}, with higher correlation to edge-based than information-content-based alternatives. It also achieves competitive downstream performance, suggesting that it captures clinically meaningful ontology structure while remaining simple to compute.

\subsection{End-to-End Evaluation of \method}
\label{sec:full_pipeline_results}
In this section, we evaluate the complete \method\ pipeline end-to-end. Section \ref{sec:end_to_end_performance} compares \method\ against unimodal and multimodal baselines, demonstrating state-of-the-art performance among methods that use only vital signs at inference. Section \ref{sec:ablation_study_2} validates the two-stage design through ablations, showing that each stage contributes meaningfully to the overall performance.

\subsubsection{Performance Evaluations}
\label{sec:end_to_end_performance}

\textbf{Baselines.}
We compare our framework against a range of baselines proposed in the literature. We first consider \textbf{unimodal baselines} trained solely on vital signs. Specifically, we use a \textit{Transformer-based architecture} \citep{vaswani2023attentionneed}, which leverages self-attention to model temporal dependencies and has been shown to outperform LSTMs on MIMIC benchmarks \citep{song2018attend}. We also compare against \textbf{multimodal supervision baselines} that incorporate clinical notes during training only. These include: (a) \textit{Cross-Modal Contrastive}, which aligns notes and vitals using a CLIP-style objective and then fine-tunes the vitals encoder \citep{ketabi2025bridging}; (b) \textit{CKLE}, which adds flat diagnosis similarity as an additional regularizer on top of the CLIP objective \citep{ding2024distilling}; and 
(c) \textit{Synthetic Notes Augmentation}, which trains a note generator and augments the vitals predictor with synthesized notes, a strategy widely used in medical imaging \citep{chartsias2017multimodal, yu20183d}.\\
\newline
\textbf{Results.}
Table~\ref{tab:performance} summarizes the results. Overall, multimodal training outperforms unimodal baselines, demonstrating the benefit of incorporating clinical notes during training. Among multimodal baselines, synthetic note augmentation performs worse, suggesting that techniques effective in medical imaging may not transfer to clinical notes due to their noise and variability. Cross-modal contrastive training is competitive with CKLE, whereas \method\ consistently achieves the best performance across tasks and label fractions.

\begin{table}[t]
\centering
\normalsize 
\setlength{\tabcolsep}{6pt} 
\renewcommand{\arraystretch}{1.15}

\begin{adjustbox}{max width=\linewidth}
\begin{tabular}{l c cc cc}
\toprule
& & \multicolumn{2}{c}{\textbf{In-Hospital Mortality}} & \multicolumn{2}{c}{\textbf{Length of Stay}} \\
\cmidrule(lr){3-4}\cmidrule(lr){5-6}
\textbf{Method} & \textbf{Labels} & \textbf{AUROC} & \textbf{AUPRC} & \textbf{AUROC} & \textbf{AUPRC} \\
\midrule

\multicolumn{6}{l}{\textbf{Unimodal}}\\
\midrule
\multirow{2}{*}{Transformer}
          & 50\%  & 0.763 (0.736--0.788) & 0.328 (0.280--0.381) & \second{0.673} (0.661--0.684) & \second{0.165} (0.160--0.175) \\
          & 100\% & \second{0.788} (0.762--0.811) & 0.372 (0.320--0.422) & 0.676 (0.664--0.688) & 0.168 (0.164--0.179) \\
\midrule

\multicolumn{6}{l}{\textbf{Multimodal (training time supervision)}}\\
\midrule
\multirow{2}{*}{Cross-Modal Contrastive}
          & 50\%  & 0.763 (0.736--0.788) & \second{0.345} (0.295--0.401) & 0.666 (0.654--0.678) & 0.164 (0.159--0.175) \\
          & 100\% & 0.773 (0.746--0.798) & \second{0.374} (0.326--0.425) & 0.673 (0.662--0.684) & \second{0.168} (0.164--0.179) \\
\midrule

\multirow{2}{*}{CKLE}
          & 50\%  & \second{0.767} (0.741--0.791) & 0.344 (0.297--0.397) & 0.672 (0.661--0.685) & 0.163 (0.159--0.174) \\
          & 100\% & 0.781 (0.754--0.805) & 0.361 (0.313--0.414) & \second{0.678} (0.667--0.690) & 0.167 (0.163--0.177) \\
\midrule

\multirow{2}{*}{Synthetic Notes Augmentation}
          & 50\%  & 0.759 (0.731--0.786) & 0.334 (0.287--0.386) & 0.665 (0.653--0.677) & 0.161 (0.157--0.170) \\
          & 100\% & 0.774 (0.749--0.798) & 0.359 (0.310--0.412) & 0.675 (0.663--0.687) & 0.165 (0.161--0.176) \\
\midrule
\multirow{2}{*}{\method\ (Ours)}
          & 50\%  & \best{0.776} (0.751--0.800) & \best{0.361} (0.311--0.414) & \best{0.676} (0.662--0.688) & \best{0.169} (0.164--0.180) \\
          & 100\% & \best{0.793} (0.769--0.816) & \best{0.385} (0.333--0.439) & \best{0.682} (0.670--0.695) & \best{0.175} (0.171--0.186) \\
\bottomrule
\end{tabular}
\end{adjustbox}
\caption{Performance comparison between \method\ and baseline methods. \method\ consistently outperforms the unimodal baseline and other training time multimodal supervision baselines. Best is \textbf{bold}, second-best is \underline{underlined}.}
\label{tab:performance}
\end{table}

\subsubsection{Ablation Studies}
\label{sec:ablation_study_2}
\textbf{Effect of Stage~1 pretraining.} 
We first compare \method\ with student models initialized from the Stage~1 pretrained encoder against those trained from scratch (Table~\ref{tab:stage1_pretraining}). Across both tasks and label fractions, pretrained initialization consistently improves performance. These results highlight the benefits of Ontology-Aware contrastive pretraining, suggesting that it learns representations that transfer effectively to downstream distillation and improve final performance.

\vspace{0.1in}
\begin{table}[H]
\centering
\footnotesize
\setlength{\tabcolsep}{4pt}
\renewcommand{\arraystretch}{1.1}
\begin{adjustbox}{max width=\linewidth}
\begin{tabular}{llcccc}
\toprule
& & \multicolumn{2}{c}{\textbf{50\% Labels}} & \multicolumn{2}{c}{\textbf{100\% Labels}} \\
\cmidrule(lr){3-4} \cmidrule(lr){5-6}
\textbf{Task} & \textbf{Initialization} & \textbf{AUROC} & \textbf{AUPRC} & \textbf{AUROC} & \textbf{AUPRC} \\
\midrule
\multirow{2}{*}{In-Hospital Mortality} & Scratch    & 0.766 & 0.334 & 0.791 & 0.379 \\
      & Pretrained & \textbf{0.776} & \textbf{0.361} & \textbf{0.793} & \textbf{0.385} \\
\midrule
\multirow{2}{*}{Length of Stay} & Scratch    & 0.672 & 0.162 & 0.680 & 0.172 \\
      & Pretrained & \textbf{0.676} & \textbf{0.169} & \textbf{0.682} & \textbf{0.175} \\
\bottomrule
\end{tabular}
\end{adjustbox}
\caption{Effect of Stage~1 pretraining on \method\ performance.}
\label{tab:stage1_pretraining}
\end{table}

\begin{figure}[H]
    \centering
    \includegraphics[width=\linewidth]{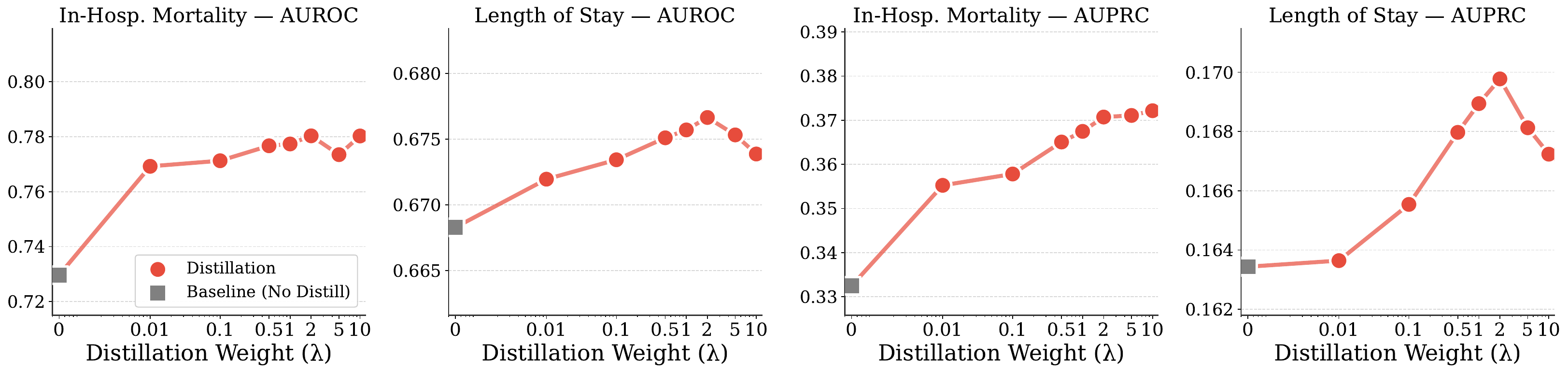}
    \vspace{-0.1in}
    \caption{Effect of $\lambda_{\text{distill}}$ on model performance, evaluated using 50\% labeled training data. Across tasks, distillation consistently improves over the no-distillation baseline. Similar trends are observed across other labeled-data fractions.}
    \label{fig:lambda_ablation}
\end{figure}

\vspace{0.1in}
\noindent\textbf{Effect of Stage~2 distillation.} We further highlight~the~benefits of distillation from clinical notes by varying the distillation weight $\lambda_{\text{distill}}$ (Figure~\ref{fig:lambda_ablation}). Across both tasks and metrics, enabling distillation ($\lambda_{\text{distill}}>0$) consistently improves over the no-distillation baseline ($\lambda_{\text{distill}}=0$). This highlights that teacher guidance from clinical notes provides complementary signal beyond supervised labels alone. Performance is strongest at moderate-to-large $\lambda_{\text{distill}}$, suggesting that distillation is most effective when soft targets contribute meaningfully to training without overwhelming the downstream objective.

\vspace{0.1in}

\section{Discussion} \label{sec:discussion}
We propose \method, a two-stage framework that improves downstream ICU risk performance on physiological time-series data. By leveraging the hierarchical structure of the ICD ontology, our approach derives more informative patient similarity measures and incorporates them into contrastive pretraining to better shape the embedding space. Building on the pretrained representations, we further demonstrate that cross-modal knowledge distillation from clinical notes provides complementary supervision, yielding additional performance gains while enabling deployment with real-time vital signs alone.

This work has several strengths. First, the framework demonstrates consistent gains across multiple tasks, label fractions, and observation horizons. We further conduct paired bootstrap testing to assess statistical significance. In the low-label setting (Table~\ref{tab:linear_prob}), OC-Distill significantly outperforms both baselines on mortality across all label fractions and metrics (all $p < 0.05$), and on length-of-stay in 5 of 6 AUROC comparisons. For the full pipeline evaluation (Table~\ref{tab:performance}), significant gains persist at the 50\% label fraction over \textit{Cross-Modal Contrastive} and \textit{Synthetic Notes Augmentation}. These results confirm significant gains, especially in the low-label regime where clinical data are often limited. Second, unlike large pretrained EHR models that often require rich inputs and substantial compute at deployment, the student model operates on vital signs alone at inference, making OC-Distill lightweight and practical for real-time ICU monitoring. Third, the ontology-aware reweighting scheme is loss-agnostic and can in principle be applied to other contrastive objectives beyond NT-Xent \citep{khosla2020supervised}, making it broadly applicable to other contrastive pretraining settings. Lastly, while we demonstrate the distillation stage with clinical notes, the teacher-student framework is modality-agnostic and naturally extends to other modalities or combinations of modalities.

This study also has its own limitations. First, our patient similarity measure depends on ICD coding, which may not fully capture clinical similarity due to noise, missing codes, or systematic biases such as upcoding \citep{silverman2004medicare, coustasse2021upcoding}. Future work could incorporate additional information such as laboratory measurements, medications, or procedures to define more robust notions of patient similarity. Second, our framework currently relies solely on clinical notes as the complementary modality. Other modalities such as medical imaging or structured EHR variables may capture complementary information not present in notes \citep{fu2021fast, shen2021labeling, hayat2022medfuse, wu2025bridging, zhang2025developing}, and exploring richer multimodal supervision remains an important direction for future work \citep{chen2025multimodal, zeng2025janusvln, zhu2026dual}. Third, our evaluation focuses on ICU vital signs, and it remains an open question whether the framework generalizes to other clinical time series and physiological waveform data \citep{tang2024advancing, oh2025multi, zhang2025machine}. Lastly, while \method\ improves predictive performance, our current analysis does not fully characterize which clinical concepts are transferred from clinical notes to vital-sign representations during distillation. Future work could further investigate which note-derived clinical concepts drive predictive performance through interpretability methods \citep{lundberg2017unified, singh2024rethinking, liang2025local, han2026interpretable}.

\acks{This work was supported by funds from Samsung Electronics Co., Ltd.}


\bibliography{references}

\appendix
\onecolumn
\section{Teacher Summary Fraction Ablation}\label{apd:teacher_summary_frac}

Our experiments use GPT-4o summaries as additional context when training the teacher model. In this appendix, we ablate the fraction of summaries included in the teacher inputs, and report teacher model performance averaged over the 50\% and 100\% label fractions. Table~\ref{tab:agg_p_summary_48h} summarizes the results. We observe that a 50\% summary fraction consistently yields the best performance, suggesting that incorporating LLM-generated summaries during training provides additional signal without overwhelming the original inputs.


\begin{table}[h]
\centering
\footnotesize
\setlength{\tabcolsep}{6pt}
\renewcommand{\arraystretch}{1.15}

\begin{adjustbox}{max width=\columnwidth}
\begin{tabular}{l c cc}
\toprule
\textbf{Task} & \textbf{Fraction of LLM Summary} & \textbf{AUROC} & \textbf{AUPRC} \\
\midrule

\multirow{3}{*}{In-Hospital Mortality}
& 0\%  & 0.872 (0.852--0.890) & 0.512 (0.459--0.565) \\
& 50\% & \best{0.879} (0.859--0.897) & \best{0.527} (0.473--0.579) \\
& 100\% & 0.856 (0.836--0.874) & 0.438 (0.386--0.495) \\
\midrule
\multirow{3}{*}{Length of Stay}
& 0\% & 0.696 (0.684--0.707) & 0.180 (0.175--0.191) \\
& 50\% & \best{0.704} (0.693--0.716) & \best{0.185} (0.180--0.198) \\
& 100\% & 0.670 (0.658--0.682) & 0.162 (0.159--0.171) \\

\bottomrule
\end{tabular}
\end{adjustbox}

\caption{Teacher model performance with varying fractions of GPT-4o summaries, averaged across the 50\% and 100\% label fractions. Best is \textbf{bold}.}
\label{tab:agg_p_summary_48h}
\end{table}

\vspace{-0.05 in}

\section{Teacher Calibration Analysis}
\label{app:teacher_calibration}

In this section, we evaluate whether incorporating clinical notes improves the calibration of the teacher's soft probabilities, further motivating the use of soft-logit distillation. As shown in Table~\ref{tab:teacher_calibration}, the multimodal teacher achieves lower expected calibration error (ECE) and Brier score than the note-free teacher across both tasks. This suggests that the teacher's soft probabilities provide additional supervision signals enriched by clinical notes that can be distilled to the student.

\begin{table}[h]
\centering
\footnotesize
\setlength{\tabcolsep}{6pt}
\renewcommand{\arraystretch}{1.15}
\begin{adjustbox}{max width=\columnwidth}
\begin{tabular}{l cc cc}
\toprule
\multirow{2}{*}{\textbf{Teacher}} & \multicolumn{2}{c}{\textbf{In-Hospital Mortality}} & \multicolumn{2}{c}{\textbf{Length of Stay}} \\
\cmidrule(lr){2-3} \cmidrule(lr){4-5}
& \textbf{ECE} & \textbf{Brier} & \textbf{ECE} & \textbf{Brier} \\
\midrule
Note-free & 0.062 & 0.171 & 0.045 & 0.751 \\
With notes & \best{0.043} & \best{0.097} & \best{0.029} & \best{0.728} \\
\bottomrule
\end{tabular}
\end{adjustbox}
\caption{Teacher calibration with and without clinical notes.}
\label{tab:teacher_calibration}
\end{table}

\vspace{-0.05 in}

\section{Supplementary Results with Longer Observation Horizons}\label{apd:first}

The main paper reports results using an observation horizon of $T=48$ hours. In this appendix, we evaluate the robustness of the method when training with longer horizons, $T\in\{72,96\}$ hours. Tables~\ref{tab:linear_prob_72h} and~\ref{tab:linear_prob_96h} present linear evaluation results for contrastive pretraining, while Tables~\ref{tab:performance_72h} and~\ref{tab:performance_96h} report full fine-tuning performance with knowledge distillation. The results at $T\in\{72,96\}$ hours follow the same trends as the main paper. Our method performs strongly across tasks and stays competitive across all settings.


\begin{table}[!t]
\centering
\footnotesize
\setlength{\tabcolsep}{6pt}
\renewcommand{\arraystretch}{1.15}

\begin{adjustbox}{max width=\linewidth}
\begin{tabular}{l c cc cc}
\toprule
& & \multicolumn{2}{c}{\textbf{In-Hospital Mortality}} & \multicolumn{2}{c}{\textbf{Length of Stay}} \\
\cmidrule(lr){3-4}\cmidrule(lr){5-6}
\textbf{Methods} & \textbf{Labels} & \textbf{AUROC} & \textbf{AUPRC} & \textbf{AUROC} & \textbf{AUPRC} \\
\midrule

\multirow{3}{*}{SimCLR}
          & 1\%  & 0.641 (0.606--0.677) & 0.228 (0.194--0.272) & 0.576 (0.560--0.591) & 0.130 (0.126--0.140) \\
          & 5\%  & \second{0.675 (0.643--0.707)} & \second{0.265 (0.223--0.314)} & 0.588 (0.570--0.606) & \second{0.145 (0.139--0.156)} \\
          & 10\% & \second{0.736 (0.707--0.763)} & \best{0.313 (0.267--0.369)} & 0.602 (0.586--0.619) & \second{0.149 (0.144--0.161)} \\
\midrule

\multirow{3}{*}{Flat Diagnosis CL}
          & 1\%  & \second{0.662 (0.626--0.700)} & \second{0.254 (0.215--0.305)} & \second{0.579 (0.563--0.594)} & \second{0.132 (0.128--0.141)} \\
          & 5\%  & 0.674 (0.642--0.706) & 0.261 (0.220--0.312) & \second{0.594 (0.577--0.612)} & \best{0.145 (0.140--0.157)} \\
          & 10\% & 0.731 (0.700--0.760) & \best{0.313 (0.267--0.369)} & \second{0.610 (0.593--0.627)} & 0.149 (0.143--0.160) \\
\midrule

\multirow{3}{*}{Ontology-Aware CL (Ours)}
          & 1\%  & \best{0.681 (0.649--0.715)} & \best{0.264 (0.225--0.314)} & \best{0.585 (0.569--0.600)} & \best{0.134 (0.130--0.145)} \\
          & 5\%  & \best{0.694 (0.663--0.725)} & \best{0.274 (0.231--0.326)} & \best{0.600 (0.582--0.617)} & 0.144 (0.139--0.155) \\
          & 10\% & \best{0.738 (0.709--0.765)} & \best{0.313 (0.267--0.369)} & \best{0.616 (0.599--0.631)} & \best{0.150 (0.145--0.162)} \\
\bottomrule
\end{tabular}
\end{adjustbox}

\caption{Linear evaluation of contrastive pretraining methods with a 72-hour observation horizon. Best is \textbf{bold}, second-best is \underline{underlined}.}
\label{tab:linear_prob_72h}
\end{table}


\begin{table}[!t]
\centering
\footnotesize 
\setlength{\tabcolsep}{6pt} 
\renewcommand{\arraystretch}{1.15}

\begin{adjustbox}{max width=\linewidth}
\begin{tabular}{l c cc cc}
\toprule
& & \multicolumn{2}{c}{\textbf{In-Hospital Mortality}} & \multicolumn{2}{c}{\textbf{Length of Stay}} \\
\cmidrule(lr){3-4}\cmidrule(lr){5-6}
\textbf{Methods} & \textbf{Labels} & \textbf{AUROC} & \textbf{AUPRC} & \textbf{AUROC} & \textbf{AUPRC} \\
\midrule

\multirow{3}{*}{SimCLR}
          & 1\%  & \second{0.619 (0.580--0.661)} & \second{0.238 (0.199--0.290)} & 0.563 (0.545--0.579) & 0.128 (0.123--0.140) \\
          & 5\%  & \second{0.676 (0.635--0.712)} & \second{0.282 (0.235--0.340)} & \second{0.578 (0.560--0.596)} & 0.143 (0.137--0.157) \\
          & 10\% & \second{0.704 (0.667--0.739)} & \best{0.310 (0.258--0.368)} & \second{0.575 (0.558--0.593)} & 0.144 (0.138--0.159) \\
\midrule

\multirow{3}{*}{Flat Diagnosis CL}
          & 1\%  & 0.618 (0.579--0.660) & \best{0.239 (0.199--0.293)} & \second{0.564 (0.548--0.581)} & \second{0.128 (0.124--0.141)} \\
          & 5\%  & 0.670 (0.629--0.708) & 0.278 (0.233--0.337) & 0.577 (0.559--0.596) & \second{0.143 (0.137--0.157)} \\
          & 10\% & 0.698 (0.661--0.734) & 0.303 (0.254--0.364) & 0.575 (0.557--0.592) & \second{0.146 (0.139--0.161)} \\
\midrule

\multirow{3}{*}{Ontology-Aware CL (Ours)}
          & 1\%  & \best{0.621 (0.582--0.662)} & 0.237 (0.200--0.289) & \best{0.565 (0.548--0.582)} & \best{0.130 (0.126--0.144)} \\
          & 5\%  & \best{0.685 (0.646--0.721)} & \best{0.289 (0.243--0.343)} & \best{0.584 (0.566--0.602)} & \best{0.147 (0.141--0.162)} \\
          & 10\% & \best{0.714 (0.678--0.746)} & \second{0.309 (0.260--0.367)} & \best{0.587 (0.570--0.606)} & \best{0.148 (0.142--0.163)} \\
\bottomrule
\end{tabular}
\end{adjustbox}

\caption{Linear evaluation of contrastive pretraining methods with a 96-hour observation horizon. Best is \textbf{bold}, second-best is \underline{underlined}.}
\label{tab:linear_prob_96h}
\end{table}


\begin{table}[!t]
\centering
\footnotesize
\setlength{\tabcolsep}{6pt}
\renewcommand{\arraystretch}{1.15}
\begin{adjustbox}{max width=\linewidth}
\begin{tabular}{l c cc cc}
\toprule
& & \multicolumn{2}{c}{\textbf{In-Hospital Mortality}} & \multicolumn{2}{c}{\textbf{Length of Stay}} \\
\cmidrule(lr){3-4}\cmidrule(lr){5-6}
\textbf{Method} & \textbf{Labels} & \textbf{AUROC} & \textbf{AUPRC} & \textbf{AUROC} & \textbf{AUPRC} \\
\midrule
\multirow{2}{*}{Transformer}
          & 50\%  & 0.719 (0.689--0.748) & 0.285 (0.246--0.334) & 0.649 (0.634--0.662) & 0.162 (0.156--0.173) \\
          & 100\% & 0.725 (0.697--0.754) & 0.293 (0.251--0.348) & \second{0.659} (0.645--0.672) & \second{0.169} (0.163--0.181) \\
\midrule
\multirow{2}{*}{Cross-Modal Contrastive}
          & 50\%  & \second{0.725} (0.697--0.757) & \second{0.312} (0.267--0.370) & 0.641 (0.627--0.656) & \second{0.162} (0.157--0.175) \\
          & 100\% & \second{0.737} (0.710--0.767) & 0.320 (0.275--0.382) & 0.649 (0.634--0.663) & 0.165 (0.159--0.177) \\
\midrule

\multirow{2}{*}{CKLE}
          & 50\%  & 0.714 (0.682--0.744) & 0.279 (0.240--0.326) & \second{0.652} (0.638--0.665) & 0.161 (0.157--0.172) \\
          & 100\% & 0.733 (0.704--0.762) & 0.320 (0.275--0.374) & \best{0.661} (0.648--0.674) & \best{0.170} (0.165--0.183) \\
\midrule

\multirow{2}{*}{Synthetic Notes Augmentation}
          & 50\%  & 0.720 (0.691--0.748) & 0.281 (0.243--0.332) & 0.639 (0.625--0.653) & 0.157 (0.152--0.168) \\
          & 100\% & 0.737 (0.706--0.767) & \second{0.326} (0.279--0.384) & 0.654 (0.640--0.668) & 0.167 (0.162--0.180) \\
\midrule

\multirow{2}{*}{\method\ (Ours)}
          & 50\%  & \best{0.751} (0.723--0.780) & \best{0.350} (0.304--0.411) & \best{0.654} (0.641--0.667) & \best{0.164} (0.159--0.175) \\
          & 100\% & \best{0.770} (0.742--0.798) & \best{0.372} (0.323--0.433) & 0.657 (0.643--0.671) & 0.166 (0.162--0.178) \\
\bottomrule
\end{tabular}
\end{adjustbox}
\caption{Performance comparison between \method\ and baseline methods with a 72-hour observation horizon. Best is \textbf{bold}, second-best is \underline{underlined}.}
\label{tab:performance_72h}
\end{table}


\begin{table}[h]
\centering
\footnotesize
\setlength{\tabcolsep}{6pt}
\renewcommand{\arraystretch}{1.15}

\begin{adjustbox}{max width=\linewidth}
\begin{tabular}{l c cc cc}
\toprule
& & \multicolumn{2}{c}{\textbf{In-Hospital Mortality}} & \multicolumn{2}{c}{\textbf{Length of Stay}} \\
\cmidrule(lr){3-4}\cmidrule(lr){5-6}
\textbf{Method} & \textbf{Labels} & \textbf{AUROC} & \textbf{AUPRC} & \textbf{AUROC} & \textbf{AUPRC} \\
\midrule
\multirow{2}{*}{Transformer}
          & 50\%  & 0.651 (0.613--0.686) & 0.251 (0.213--0.305) & 0.627 (0.609--0.644) & \second{0.160} (0.153--0.173) \\
          & 100\% & 0.693 (0.655--0.728) & 0.309 (0.256--0.368) & 0.638 (0.621--0.654) & 0.164 (0.157--0.179) \\
\midrule
\multirow{2}{*}{Cross-Modal Contrastive}
          & 50\%  & 0.533 (0.491--0.572) & 0.175 (0.148--0.209) & 0.561 (0.542--0.579) & 0.129 (0.125--0.141) \\
          & 100\% & 0.573 (0.532--0.611) & 0.186 (0.160--0.225) & 0.630 (0.611--0.647) & 0.159 (0.151--0.176) \\
\midrule

\multirow{2}{*}{CKLE}
          & 50\%  & 0.663 (0.625--0.699) & \second{0.279} (0.234--0.341) & \second{0.629} (0.612--0.645) & 0.155 (0.150--0.167) \\
          & 100\% & 0.701 (0.665--0.737) & 0.307 (0.258--0.363) & \second{0.639} (0.621--0.655) & \second{0.165} (0.158--0.178) \\
\midrule

\multirow{2}{*}{Synthetic Notes Augmentation}
          & 50\%  & \second{0.675} (0.637--0.711) & 0.271 (0.229--0.327) & 0.628 (0.611--0.644) & 0.158 (0.152--0.170) \\
          & 100\% & \second{0.736} (0.699--0.769) & \second{0.335} (0.283--0.398) & 0.631 (0.614--0.648) & 0.161 (0.155--0.178) \\
\midrule

\multirow{2}{*}{\method\ (Ours)}
          & 50\%  & \best{0.731} (0.694--0.764) & \best{0.358} (0.302--0.419) & \best{0.641} (0.623--0.656) & \best{0.165} (0.159--0.180) \\
          & 100\% & \best{0.742} (0.709--0.775) & \best{0.350} (0.295--0.413) & \best{0.639} (0.621--0.654) & \best{0.165} (0.159--0.178) \\
\bottomrule
\end{tabular}
\end{adjustbox}
\caption{Performance comparison between \method\ and baseline methods with a 96-hour observation horizon. Best is \textbf{bold}, second-best is \underline{underlined}}
\label{tab:performance_96h}
\end{table}

\section{Sensitivity to the choice of $K$ in neighbor analysis}
\label{apd:knn_sensitivity}

We repeated the embedding-space neighbor analysis with $K=1$ and $K=3$ in addition to the main-text setting of $K=5$. As shown in Table~\ref{tab:knn_sensitivity}, the conclusion is stable across all choices of $K$: nearest-neighbor pairs consistently exhibit higher diagnosis similarity than randomly sampled pairs, and all differences remain statistically significant.

\begin{table}[h]
\centering
\begin{tabular}{ccccc}
\toprule
$K$ & KNN Mean & Random Mean & Effect Size $r$ & Mann--Whitney $p$ \\
\midrule
1 & 0.251 & 0.200 & 0.214 & $< 0.0001$ \\
3 & 0.248 & 0.200 & 0.205 & $< 0.0001$ \\
5 & 0.246 & 0.199 & 0.198 & $< 0.0001$ \\
\bottomrule
\end{tabular}
\caption{Sensitivity of the embedding-space similarity analysis to the choice of neighborhood size $K$.}
\label{tab:knn_sensitivity}
\end{table}

\vspace{-.15 in}

\section{Comparison with Established Ontology Similarity Metrics}
\label{app:ontology_similarity_metrics}

We compare our Jaccard path-overlap similarity against established ontology similarity measures commonly used in the biomedical ontology literature, including edge-based and information-content-based metrics \citep{rada1989development, resnik1995using, jiang1997semantic}. We first assess whether our simple ICD path-overlap metric captures a structure similar to that of established ontology measures and then evaluate whether the choice of ontology similarity metric affects downstream performance.

Table~\ref{tab:ontology_metric_correlation} reports Spearman correlations between our similarity metric and these established alternatives. Results show that our metric is highly correlated with all three measures, with the strongest correlation to the edge-based metric as expected. Table~\ref{tab:ontology_metric_downstream} reports linear evaluation performance. Our simple metric achieves competitive or best performance across label fractions, suggesting that a simple Jaccard metric captures the clinically relevant structure needed for contrastive pretraining while avoiding additional complexity.

\begin{table}[h]
\centering
\footnotesize
\setlength{\tabcolsep}{6pt}
\renewcommand{\arraystretch}{1.1}
\begin{tabular}{lc}
\toprule
\textbf{Metric} & \textbf{Spearman $\rho$} \\
\midrule
Rada~\citep{rada1989development} & 0.950 \\
Jiang--Conrath~\citep{jiang1997semantic} & 0.934 \\
Resnik~\citep{resnik1995using} & 0.840 \\
\bottomrule
\end{tabular}
\caption{Correlation between our ICD path-overlap similarity and established ontology similarity metrics.}
\label{tab:ontology_metric_correlation}
\end{table}

\vspace{-.1in}

\begin{table}[h]
\centering
\footnotesize
\setlength{\tabcolsep}{4pt}
\renewcommand{\arraystretch}{1.1}
\begin{adjustbox}{max width=\linewidth}
\begin{tabular}{l ccc ccc}
\toprule
\multirow{2}{*}{\textbf{Metric}} & \multicolumn{3}{c}{\textbf{AUROC}} & \multicolumn{3}{c}{\textbf{AUPRC}} \\
\cmidrule(lr){2-4} \cmidrule(lr){5-7}
& \textbf{1\%} & \textbf{5\%} & \textbf{10\%} & \textbf{1\%} & \textbf{5\%} & \textbf{10\%} \\
\midrule
Resnik~\citep{resnik1995using} & 0.651 & 0.735 & 0.744 & 0.210 & 0.298 & 0.305 \\
Jiang--Conrath~\citep{jiang1997semantic} & 0.669 & 0.745 & 0.755 & 0.225 & 0.311 & 0.319 \\
Rada~\citep{rada1989development} & 0.669 & \best{0.751} & \best{0.758} & 0.229 & \best{0.321} & \best{0.332} \\
Ours & \best{0.673} & 0.750 & \best{0.758} & \best{0.230} & 0.319 & 0.328 \\
\bottomrule
\end{tabular}
\end{adjustbox}
\caption{Downstream performance using different ontology similarity metrics in contrastive pretraining. Results are reported for linear evaluation under 1\%, 5\%, and 10\% labeled training data. Best results are \textbf{bold}.}
\label{tab:ontology_metric_downstream}
\end{table}

\section{Validation on MIMIC-IV}
\label{apd:mimic4}
To assess the generalizability of \method\ beyond MIMIC-III, we evaluate the student model on vital signs from MIMIC-IV \citep{johnson2023mimic}. Since our framework requires only vital signs at inference, the student model can be directly applied to MIMIC-IV without any retraining. We follow the same preprocessing pipeline and benchmark tasks as in Section~\ref{sec: experiments}. Table~\ref{tab:mimic4} reports results, where \method\ achieves strong performance across both tasks.
\begin{table}[h]
\centering
\footnotesize
\setlength{\tabcolsep}{4pt}
\renewcommand{\arraystretch}{1.1}
\begin{adjustbox}{max width=\linewidth}
\begin{tabular}{l cc cc}
\toprule
& \multicolumn{2}{c}{\textbf{In-Hospital Mortality}} & \multicolumn{2}{c}{\textbf{Length of Stay}} \\
\cmidrule(lr){2-3}\cmidrule(lr){4-5}
\textbf{Method} & \textbf{AUROC} & \textbf{AUPRC} & \textbf{AUROC} & \textbf{AUPRC} \\
\midrule
Transformer & 0.792 (0.775--0.809) & 0.368 (0.333--0.408) & 0.668 (0.658--0.676) & 0.158 (0.155--0.165) \\
Cross-Modal Contrastive & 0.792 (0.775--0.809) & 0.361 (0.326--0.399) & 0.670 (0.661--0.680) & \best{0.162 (0.158--0.170)} \\
CKLE & 0.797 (0.780--0.813) & 0.357 (0.322--0.394) & 0.668 (0.658--0.677) & 0.157 (0.153--0.163) \\
Synthetic Notes Augmentation & 0.794 (0.777--0.811) & 0.367 (0.331--0.406) & 0.665 (0.655--0.674) & 0.154 (0.151--0.161) \\
\method\ (Ours) & \best{0.813 (0.797--0.828)} & \best{0.380 (0.346--0.419)} & \best{0.672 (0.663--0.682)} & 0.158 (0.155--0.165) \\
\bottomrule
\end{tabular}
\end{adjustbox}
\caption{Validation on MIMIC-IV vital signs. \method\ demonstrates strong performance across both tasks. Best results are \textbf{bold}.}
\label{tab:mimic4}
\end{table}

\section{Validation on eICU}
\label{apd:eicu}

To further assess the generalizability of \method\ beyond MIMIC-III, we evaluate the student model on vital signs from the eICU dataset \citep{pollard2018eicu}. Tables~\ref{tab:eicu_mortality} and~\ref{tab:eicu_los} report results for in-hospital mortality and length-of-stay prediction. \method\ achieves strong performance for mortality prediction and remains competitive for length-of-stay prediction across horizons.

\begin{table}[h]
\centering
\footnotesize
\setlength{\tabcolsep}{4pt}
\renewcommand{\arraystretch}{1.1}
\begin{adjustbox}{max width=\linewidth}
\begin{tabular}{l cc cc cc}
\toprule
\multirow{2}{*}{\textbf{Method}} 
& \multicolumn{2}{c}{\textbf{48h}} 
& \multicolumn{2}{c}{\textbf{72h}} 
& \multicolumn{2}{c}{\textbf{96h}} \\
\cmidrule(lr){2-3}\cmidrule(lr){4-5}\cmidrule(lr){6-7}
& \textbf{AUROC} & \textbf{AUPRC} 
& \textbf{AUROC} & \textbf{AUPRC} 
& \textbf{AUROC} & \textbf{AUPRC} \\
\midrule
Transformer 
& 0.714 (0.708--0.720) 
& 0.262 (0.252--0.271) 
& 0.702 (0.694--0.709) 
& 0.273 (0.263--0.284) 
& 0.660 (0.650--0.669) 
& 0.255 (0.245--0.267) \\
Cross-Modal Contrastive 
& 0.707 (0.701--0.714) 
& 0.255 (0.245--0.264) 
& 0.696 (0.689--0.704) 
& 0.256 (0.246--0.267) 
& 0.591 (0.582--0.600) 
& 0.203 (0.195--0.212) \\
CKLE 
& 0.731 (0.724--0.737) 
& 0.267 (0.258--0.277) 
& 0.700 (0.693--0.707) 
& 0.265 (0.255--0.275) 
& 0.670 (0.661--0.680) 
& 0.251 (0.242--0.263) \\
Synthetic Notes Augmentation 
& 0.722 (0.716--0.728) 
& 0.266 (0.256--0.276) 
& 0.693 (0.685--0.700) 
& \best{0.275 (0.265--0.287)} 
& 0.660 (0.651--0.669) 
& 0.254 (0.244--0.265) \\
\method\ (Ours) 
& \best{0.746 (0.740--0.751)} 
& \best{0.274 (0.265--0.284)} 
& \best{0.703 (0.695--0.710)} 
& 0.273 (0.263--0.285) 
& \best{0.675 (0.666--0.685)} 
& \best{0.273 (0.263--0.286)} \\
\bottomrule
\end{tabular}
\end{adjustbox}
\caption{External validation on eICU vital signs for in-hospital mortality prediction across 48, 72, and 96 hour observation horizons. Best results are \textbf{bold}.}
\label{tab:eicu_mortality}
\end{table}

\begin{table}[h]
\centering
\footnotesize
\setlength{\tabcolsep}{4pt}
\renewcommand{\arraystretch}{1.1}
\begin{adjustbox}{max width=\linewidth}
\begin{tabular}{l cc cc cc}
\toprule
\multirow{2}{*}{\textbf{Method}} 
& \multicolumn{2}{c}{\textbf{48h}} 
& \multicolumn{2}{c}{\textbf{72h}} 
& \multicolumn{2}{c}{\textbf{96h}} \\
\cmidrule(lr){2-3}\cmidrule(lr){4-5}\cmidrule(lr){6-7}
& \textbf{AUROC} & \textbf{AUPRC} 
& \textbf{AUROC} & \textbf{AUPRC} 
& \textbf{AUROC} & \textbf{AUPRC} \\
\midrule
Transformer 
& 0.640 (0.637--0.643) 
& \best{0.143 (0.141--0.144)} 
& 0.625 (0.621--0.628) 
& 0.140 (0.139--0.142) 
& 0.608 (0.604--0.612) 
& \best{0.139 (0.137--0.141)} \\
Cross-Modal Contrastive 
& 0.640 (0.637--0.643) 
& 0.142 (0.141--0.144) 
& 0.623 (0.619--0.626) 
& 0.139 (0.138--0.141) 
& 0.577 (0.573--0.581) 
& 0.125 (0.124--0.127) \\
CKLE 
& \best{0.641 (0.638--0.644)} 
& 0.142 (0.141--0.144) 
& 0.624 (0.621--0.628) 
& 0.140 (0.139--0.142) 
& \best{0.609 (0.606--0.613)} 
& 0.138 (0.136--0.140) \\
Synthetic Notes Augmentation 
& 0.639 (0.636--0.641) 
& 0.142 (0.141--0.143) 
& 0.620 (0.616--0.623) 
& 0.139 (0.137--0.140) 
& 0.606 (0.602--0.609) 
& 0.137 (0.136--0.139) \\
\method\ (Ours) 
& 0.639 (0.636--0.642) 
& 0.141 (0.140--0.143) 
& \best{0.629 (0.626--0.633)} 
& \best{0.143 (0.141--0.145)} 
& 0.609 (0.605--0.612)
& 0.138 (0.137--0.141) \\
\bottomrule
\end{tabular}
\end{adjustbox}
\caption{External validation on eICU vital signs for length-of-stay prediction across 48, 72, and 96 hour observation horizons. Best results are \textbf{bold}.}
\label{tab:eicu_los}
\end{table}

\end{document}